\documentclass{article}

\usepackage{PRIMEarxiv}

\usepackage[utf8]{inputenc} 
\usepackage[T1]{fontenc}    
\usepackage{hyperref}       
\usepackage{url}            
\usepackage{booktabs}       
\usepackage{amsfonts}       
\usepackage{nicefrac}       
\usepackage{microtype}      
\usepackage{lipsum}
\usepackage{fancyhdr}       
\usepackage{graphicx}       
\graphicspath{{media/}}     

\usepackage{graphicx}%
\usepackage{multirow}%
\usepackage{amsmath,amssymb,amsfonts}%
\usepackage{amsthm}%
\usepackage{mathrsfs}%
\usepackage[title]{appendix}%
\usepackage{xcolor}%
\usepackage{textcomp}%
\usepackage{manyfoot}%
\usepackage{booktabs}%
\usepackage{algorithm}%
\usepackage{algorithmicx}%
\usepackage{algpseudocode}%
\usepackage{listings}%
\usepackage{pythonhighlight}
\usepackage{tikz}
\usetikzlibrary{shapes.geometric, arrows.meta, positioning}
\usepackage{algorithm}
\usepackage{algpseudocode}
\usepackage{orcidlink,thumbpdf,lmodern}
\usepackage{xcolor}
\usepackage{amssymb}
\usepackage{scalerel}
\usepackage{stackengine}
\usepackage{pgf}
\newcounter{iloop}
\newcommand\openbigstar[1][0.7]{%
  \scalerel*{%
    \stackinset{c}{-.125pt}{c}{}{\scalebox{#1}{\color{white}{$\bigstar$}}}{%
      $\bigstar$}%
  }{\bigstar}
}
\newcommand{\Stars}[1]{\ensuremath{%
\pgfmathtruncatemacro{\imax}{ifthenelse(int(#1)==#1,#1-1,#1)}%
\pgfmathsetmacro{\xrest}{0.9*(1-#1+\imax)}%
\setcounter{iloop}{0}%
\loop\stepcounter{iloop}\ifnum\value{iloop}<\the\numexpr\imax+1
\bigstar\repeat
\openbigstar[\xrest]%
\setcounter{iloop}{0}%
\loop\stepcounter{iloop}\ifnum\value{iloop}<\the\numexpr5-\imax\relax
\openbigstar[.9]\repeat}}

\tikzstyle{db} = [cylinder, shape border rotate=90, aspect=0.25, draw, minimum height=2cm, minimum width=1cm, fill=blue!10]
\tikzstyle{file} = [rectangle, draw, fill=gray!10, minimum width=2.5cm, minimum height=1cm]
\tikzstyle{llm} = [rectangle, draw, rounded corners, fill=orange!20, minimum width=2.5cm, minimum height=1cm]
\tikzstyle{arrow} = [thick, ->, >=stealth]
\tikzstyle{danger} = [red, font=\footnotesize]
\tikzstyle{safe} = [green!60!black, font=\footnotesize]

\title{Improving LLMs with a knowledge from databases
}

\author{
  Petr {M\'{a}\v{s}a~}\orcidlink{0000-0002-1744-034X} \\
  Department of Information and Knowledge Engineering \\
  Prague University of Economics and Business \\
  Prague, Czechia\\
  \texttt{masa@petrmasa.com} 
}

\begin{document}
\maketitle

\begin{abstract}
Large language models (LLMs) are achieving significant progress almost every moment now. Many advanced techniques have been introduced and widely accepted, like retrieval-augmentation generation (RAG), agents, and tools. 
Tools can query the database to answer questions from structured data files or perform groupings or other statistics. This unlocks huge opportunities, such as it can answer any question, but also poses threats, such as safety, because there is no control over the commands that are created. We would like to discuss whether we can create a new method that improves answers based on dataset/database via some interpretable ML methods, namely enhanced association rules. The advantage would be if the method can be also used in some safe technique like RAG.

Association rules have a sound history. Since the introduction of CN2 and aproiri, many enhancements have been made. In parallel, enhanced association rules have been introduced and evolved over the last 40 years. The general problem is typically that there are too many rules. There are some techniques for handling it, but when LLM emerged, it turned out to be the best use case for the RAG technique for LLMs. We proposed a method that generates a ruleset based on defined knowledge patterns, then converts rules into text form via a rule-to-text converter, and includes the result as an RAG into LLM. We compared this method with ChatGPT (even with using agents) and we have discovered a significant improvement in answering questions based on the dataset. We have also tried several strategies how much rules to generate. We found this improvement interesting. Moreover, it can also be improved in many ways as future work, like incorporating other patterns, the use of rule mining as an agent, and many others.\\{\bf Repository: \url{https://www.github.com/petrmasa/clmrag}}
\end{abstract}


\section{Introduction}
Large language models (LLMs) made a rapid growth over the last few years. They can perform many tasks and use cases. In fact, use cases are being refined for production use. One of the use cases of LLMs is to answer questions based on data. This is typically solved via agents that create an SQL query or other analytical procedure, run this code against the database, and based on the result, form the answer. This way is very powerful as users can have answers for data questions quickly and without a knowledge of data languages, but it currently requires skills in prompt engineering, and it has some risks of inappropriate queries built by automated way (syntactically incorrect, not powerful enough or even risk of harmful code generation that could lead to damage of data or system or overloading of system). We observed that today's LLMs are great in easy data tasks like preparing profiles, groupings, or forming a set of SQL calls based on text-based questions. These techniques are fine and can be improved to provide answers not solely based on data, but based also on knowledge discovered from data. For that reason we chose one of the interpretable methods -- enhanced association rules. These rules are an interpretable method for knowledge discovery from databases.

Enhanced association rule in its simplest form is a rule $A\Rightarrow S$, where $A$ and $S$ are typically conjunctions of literals, where a literal is the condition on an attribute and its values. For an easy interpretation, the method is designed for categorical variables, but it is not a problem as continuous variables can be easily converted to categorical (ordinal) ones, and enhanced association rules also support category joining, so the binning can be done automatically, and a number of categories does not matter too much. 

\begin{figure}[htp]
    \centering
    \includegraphics[width=0.7\textwidth]{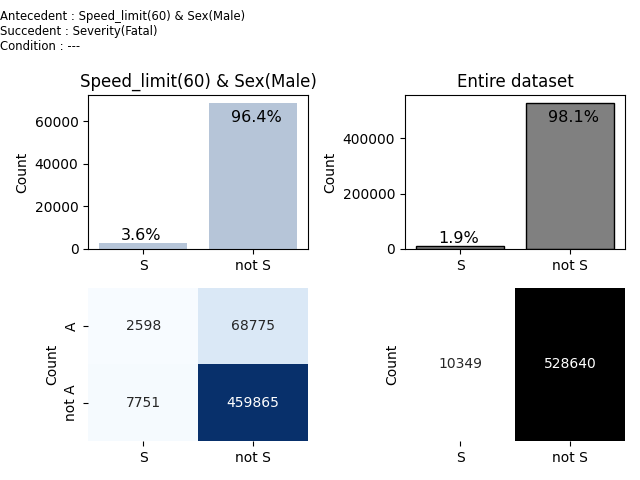}
    \caption{Severity of fatal accident is almost doubled for male drivers on speed limit 60 compared to the entire dataset.}
    \label{fig:fst_rule}
\end{figure}

We will start with an example of enhanced association rules based on the Kaggle UK accidents dataset.
This dataset contains information about UK traffic accidents, including 538,989 road accidents and their properties. 
There is 1.9\% Fatal accidents. We might be curious about which segment of drivers or circumstances there are more or fewer Fatal accidents. The example of enhanced association rule is 
$$Speed\_Limit(60) \& Sex(Male) \Rightarrow_{conf=3.6\%} Severity(Fatal),$$ i.e., that there is 3.6\% probability that the accident will be Fatal when driven by a male at speed limit 60. That is almost doubled compared with the probability of Fatal accident on the entire dataset (which is 1.9\%). The situation is shown in Figure \ref{fig:fst_rule}.
This rule is the easiest pattern of enhanced association rules called 4ft-Miner. There are also several other patterns that can extract an interesting knowledge. All these rules are easily interpretable.

Note that enhanced association rules can also be mined from a contingency (pivot) table in a spreadsheet in a small scope, so it can also be named a search on contingency tables (SCT). Pivot tables are widely used due to wide adoption of spreadsheets and an easy interpretation.

Our goal in this article will be to assess whether these rules can easily improve LLM models. We will focus on the RAG method, which is safe, and will compare the result with automated analytic agents/tools implemented in current LLMs. 

The structure of the article is the following. Section \ref{sec:foundations} introduces foundations needed for the rest of the article, next Section \ref{sec:methodology} will introduce the new method to improve LLMs with a knowledge gained from databases, Section \ref{sec:results} will show examples and comparison of the proposed method and finally Section \ref{sec:summary} summarizes the article.

\section{Foundations}
\label{sec:foundations}

\subsection{Large language models and ways to enhance them}
Large language models (LLMs) have revolutionized natural language processing. These models are typically trained on large datasets to be able to perform a wide range of tasks. As they was trained on static data, there is a risk of outdated or incomplete information. To address this challenge, Retrieval-Augmented generation (RAG) \cite{lewis2020retrieval} was introduced. RAG enhances LLMs by incorporating external information, allowing the utilization of the most recent or domain (company) specific information during response generation.

Note that we will omit parameter-efficient fine-tuning methods such as LoRA (Low-Rank Adaptation) \cite{hu2021lora} that allow LLMs to be retrained or adapted to new domains or tasks without modifying the full parameter set.

Retrieval-Augmentation generation works on the principle that the user can input their own information into the context window (the window that is assessed during the generation of the answer). This allows it to include its own data without the need to retrain the entire model. But it consumes a valuable place in the context window, which is limited. Note that these limitations are big enough for classical querying, but for including large documents the context window might not be large enough. So the principle is the following. Documents to be included in the reasoning are split into small overlapping chunks. For each of them, the embedding function is calculated, and a similarity with the question is calculated. Most similar ones are then included in the context window for LLM when querying. This principle allows for embedding relatively large sets of documents into reasoning, while the consumption of valuable space in the context window is not so big.

The next way is to use agents. This principle allows LLMs to call an external service and use its result for reasoning. An example is that based on a query, LLM can decide whether to use a company database, create an SQL statement, ask for its execution, and, based on the result, form an answer. Another way is to use analytics procedures to profile data in Python or perform some analysis.
The comparison is in Table \ref{tab:llm_comparison}.
\begin{table}[h!]
\begin{center}
\caption{Comparative Overview of LLMs, LoRA, RAG, and Agentic AI Systems}
\label{tab:llm_comparison}
\begin{footnotesize}
\centering
\begin{tabular}{|p{1.8cm}|p{2.5cm}|p{2.5cm}|p{2.5cm}|p{2.5cm}|}
\hline
\textbf{Aspect} & \textbf{LLMs} & \textbf{LLM + LoRA (Fine-Tuning)} & \textbf{RAG} & \textbf{Agentic AI} \\
\hline
Core Function & Text generation & Domain-adapted generation & Contextualized generation via retrieval & Goal-directed task execution \\
\hline
Knowledge Source & Static (pretrained corpus) & Static + domain-specific updates & External documents (real-time) & Memory, tools, APIs \\
\hline
Adaptation Method & None or full fine-tuning & Low-rank adaptation (efficient) & No model retraining required & Modular \& dynamic reasoning \\
\hline
Latency & Low & Slightly increased & Higher (due to retrieval) & High (due to orchestration) \\
\hline
Accuracy & General-purpose & Improved domain accuracy & Improved area accuracy & Task-aligned contextual responses \\
\hline
Autonomy & None & Low & Low & High \\
\hline
\end{tabular}
\end{footnotesize}
\end{center}
\end{table}

\subsection{Interpretable ML and explainable AI}
Since the introduction of classical ML methods, many advancements have been made, typically with a hunt for better performance and neglecting interpretability. According to  \cite{BARREDOARRIETA202082} and \cite{Linardatos2020ExplainableAA}, association rules remained one of the few interpretable ML techniques, along with decision trees. Regression models are sometimes denoted as interpretable and sometimes not, as when predictors are correlated, the interpretation of regression coefficients can be misleading.

Decision trees are easy to understand and are very good for predictions as they cover each case exactly once. For knowledge discovery, it is a strong tool for initial analysis, but advanced knowledge discovery is limited as it strongly depends on splits in the root and close to the root. For example, exception mining is relatively hard to achieve. It can be overcome by interactive trees where analysts try multiple splits, but it simulates enhanced association rule mining.

Next area that emerged is explainable AI \cite{XAIanreview}, that tries to explain complex models using several techniques. Some of them are outlined in \cite{BARREDOARRIETA202082}. They consists of both classical interpretable methods like association rules and decision trees as well as novel methods like SHAP \cite{shap} or a local method LIME \cite{lime}. Note that there are disputes whether to use these methods as interpretable methods can give in many cases performance results near to black box ones with a full transparency and control \cite{Rudin2019-ub}.

So we will focus on enhanced association rules because they provide a most comprehensive knowledge discovery from databases in an understandable way, with the general disadvantage of that there are too many rules that is typically solved by LLMs that are designed to process it.

\subsection{Enhanced association rules}
Enhanced association rules have a sound history.
Since the introduction of CN2 \cite{cn2} and aproiri \cite{Ag:93}, many enhancements have been made. In parallel, enhanced association rules have been introduced \cite{Ha:78}, \cite{HaHoRa}, and have evolved over the last 40 years. Main milestones are that a production-use ready implementation of enhanced association rules became available \cite{LISPMINER_10.1007/978-3-540-44999-7_25}, \cite{Simunek:2003}, and being enhanced by principles like action rules \cite{ras2000action},\cite{Dardzinska2013}, subset detection \cite{Atzmueller:2015}, and other enhancements. In the other hand, some enhancement of association rules became available as {\em arules} package \cite{Hahsler:arules} with a portation into Python \cite{hahsler_arulespy}. Most recent publication that summarizes generalized association rules at this time is summarized in \cite{CRC_GUHA}. General problem is typically that there is too many rules.
 There are some techniques how to handle it, e.g. \cite{LOD22_CLM} or \cite{MasaRauch_JIIS}, but when LLM emerged, it turns that it is a best use case for RAG technique for LLMs.

Let focus more deeply on enhanced association rules itself. They are based on mathematical logic. The easiest pattern is the {\em 4ft-Miner} (abbreviation of four-fold table Miner) that suggests that it is based on a four fold table. 
Let start with a dataset ${\cal M}$ with attributes $A_1,A_2,\dots,A_m$. Then we define a {\em literal} as expression $A(\alpha)$, where $\alpha\subsetneq \{a_1,a_2,\dots,a_k\}$, where $\{a_1,a_2,\dots,a_k\}$ is a set of all values of attribute $A$. The literal $A(\alpha)$ is true for object $o$ in dataset $\cal M$ if and only if $M[o]\in \alpha$.

Now, let define a {\em boolean attribute}. Each literal is a boolean attribute. When $\phi$ and $\psi$ are boolean attributes, then $\phi\land\psi$, $\phi\lor\psi$ and $\neg\phi$ are also boolean attributes.

Let have an example -- $A_1(1)$ is a literal and it is true if and only if the value of $A_1$ is 1 for object $o$. The literal $A_2(6,9)$ is also literal that means that $A_2$ has value either 6 or 9. Both these literals are boolean attributes. The expression $A_1(1)\lor A_2(6,9)$ is an example of a boolean attribute that is not a literal.

The rule mined by the 4ft-Miner is a rule in the form $\phi\Rightarrow \psi$, where $\phi$ and $\psi$ are boolean attributes. $\phi$ is called {\em antecedent}, or left-hand side of the rule, and $\psi$ is called a succedent, or a right hand side of the rule.

We can define a fourfold table for boolean attributes $\phi$ and $\psi$ in a dataset $M$ as a quadruple $4ft({\cal M},\phi,\psi)=(a,b,c,d)$, where $a$ is the number of records satisfying both $\phi$ and $\psi$ in a dataset $\cal M$, $b$ is a number of attributes satisfying $\phi$ but not $\psi$ etc. The situation is shown in Table \ref{fig:ff}.

\begin{table}[htp]
\caption{Fourfold table for boolean attributes $\phi$ and $\psi$.}
\label{fig:ff}
\begin{center}
\begin{tabular}{c|c|c|}
&$\psi$&$\neg\psi$\\\hline
$\phi$&a&b\\
$\neg\phi$&c&d\\\hline
    
\end{tabular}
\end{center}
\end{table}

For each rule, we can calculate {\em quantifiers} that reflects the measure of interestingness of the rule. Typical quantifiers is {\em Base} (or a support in absolute form, abbreviated as $base$) defined as $Base = a$ and confidence (denoted as $conf$). Confidence is defined as $conf=P(\psi|\phi)=\frac{a}{a+b}$. Sometimes, it is better to use the confidence in a relative form, i.e., how much the probability of the succedent increases when the antecedent is present. This measure is called the {\em lift} and it is defined like $lift=\frac{P(\psi|\phi)}{P(\psi)}=\frac{\frac{a}{a+b}}{\frac{a+c}{a+b+c+d}}$, or a very similar measure, {\em above average difference} defined as $aad=lift-1$.

The CleverMiner \cite{cleverminer} is a Python package to mine enhanced association rules. It allows easy embedding into other applications and have a rich API. We will use this package to gain knowledge from a dataset.

\subsection{Benchmarking dataset and knowledge-based answers}

As we announced, we would like to compare results based on our proposed method with existing LLMs, including agents. So we will need to look at tests or metrics that reflect this aspect. If none of them is suitable, we will need to choose some other method, even not so sophisticated.

So let summarize the most similar benchmarks (at our best knowledge and research).

\begin{enumerate}
\item {\bf TAT-QA (Tabular and Textual Question Answering)} \cite{zhu-etal-2021-tat} introduces a hybrid data from the financial domain focusing on both textual and tabular data. This is a great library, but it focuses mainly on basic operations on data, mainly based on a defined attribute, not a comprehensive scan and data massaging of a dataset.
\item {\bf Databench} \cite{databench-oses-grijalba-etal-2024-question} is a great collection of questions based on several datasets. But also it is a set of relatively simple questions based on a single attribute (like what is the oldest passenger, what is the lowest fare paid, etc.) or a combination of 2 attributes, and the questions are relatively straightforward (an answer can be obtained by a single query).
\item {\bf QRData (Quantitative reasoning data)}\cite{liu2024llmscapabledatabasedstatistical} is another benchmark that tests how good is the LLM model in statistical reasoning. Questions are closer to the knowledge that we are looking for, but again, each question is an analytical question that could be rewritten to a few lines of code (e.g., what is the proportion of category A, excluding missings; or round the result to 3 digits, test the hypothesis, etc.) and benchmark test this ability.
\item {\bf GRS-QA (Graph Reasoning-Structured Question Answering)} \cite{pahilajani2024grsqagraphreasoningstructured} is a benchmark that tests the ability of reasoning based on facts and derivates. It is a very good test of logic, it does not test discovery of structures in data.

\item {\bf EconLogicQA} \cite{quan-liu-2024-econlogicqa} also tests sequential reasoning, but it also covers economic areas where relationships are more complex.
\end{enumerate}

Note that the last two could be a good benchmark for advanced analyses on how this method works under some scenarios, like joining categories, etc. 
So we haven't found any QA benchmark to evaluate it. 

\section{Methodology}
\label{sec:methodology}
Today's mainstream tends to use agents. The ideal scenario is to answer the specific question (ideally easy one for that SQL/program code can be easily generated). This is also the way, but as enhanced association rules can bring a strong benefit for using LLM on database data (as it mines a knowledge based on a data), we will start with an easier task. Our approach will to prepare a solid knowledge base based on a dataset. So we will extract the knowledge from data, prepare it in form understandable and embeddable by LLMs and include it in LLM. 

First, we need to focus on a pattern. We start with the simple pattern defined by 4ft-Miner, that finds rules in form {\em If $A$ Then $S$} that can easily answer questions like what causes $S$, especially when $S$ is a target variable. Note that this approach is general, any other pattern can be used as well. Then, we will define a task for a mining task, run it, get the list of rules, transform rules to the text file and use this text file as an input in RAG for LLM. The schema is shown in Fig. \ref{fig:main_schema}.


\usetikzlibrary{shapes.geometric, arrows.meta, positioning}

\tikzstyle{startstop} = [rectangle, rounded corners, minimum width=3cm, minimum height=1cm,text centered, draw=black, fill=blue!10]
\tikzstyle{process} = [rectangle, minimum width=3cm, minimum height=1cm, text centered, draw=black, fill=green!10]
\tikzstyle{data} = [trapezium, trapezium left angle=70, trapezium right angle=110, minimum width=3cm, minimum height=1cm, text centered, draw=black, fill=orange!20]
\tikzstyle{arrow} = [thick,->,>=Stealth]

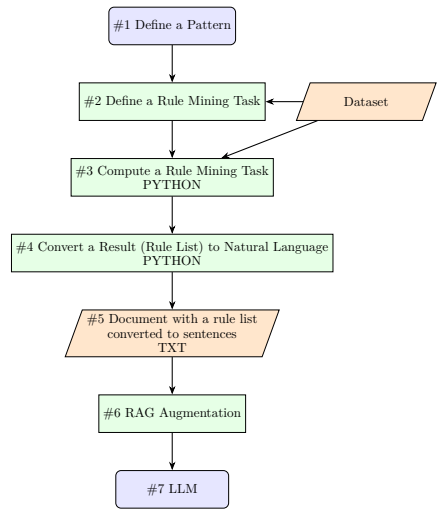
\begin{figure}[htp!]
\centering
\scalebox{.5}{
\begin{tikzpicture}[scale=2]

\node (pattern) [startstop] {\#1 Define a Pattern};
\node (mining) [process, below=of pattern] {\#2 Define a Rule Mining Task};
\node (db) [data,right=of mining] {Dataset};
\node (compute) [process, below=of mining, align=center] {\#3 Compute a Rule Mining Task\\PYTHON};
\node (nloutput) [process, below=of compute,align=center] {\#4 Convert a Result (Rule List) to Natural Language\\PYTHON};
\node (doc) [data,below=of nloutput,align=center] {\#5 Document with a rule list \\converted to sentences\\TXT};
\node (rag) [process, below=of doc] {\#6 RAG Augmentation};
\node (llm) [startstop, below=of rag] {\#7 LLM};

\draw [arrow] (pattern) -- (mining);
\draw [arrow] (mining) -- (compute);
\draw [arrow] (compute) -- (nloutput);
\draw [arrow] (db) -- (mining);
\draw [arrow] (db) -- (compute);
\draw [arrow] (nloutput) -- (doc);

\draw [arrow] (doc) -- (rag);
\draw [arrow] (rag) -- (llm);

\end{tikzpicture}}
    \caption{Proposed schema of augmenting knowledge from database into LLM}
    \label{fig:main_schema}
\end{figure}

Note that points 1--3 are parts of the standard ML (or KDD) task, points 5--7 are classical use of RAG in LLM, so we need to focus on point 4, which is the Rule-To-Text module.

Let go more deeper. We will stay at the Kaggle UK accident dataset. We will focus on a severity of accidents (that might be considered as a target variable) and look for reasons for more serious accidents (serious or fatal). Note that pattern can be anything so we can focus on much more types of analytical questions.

One more note compared to a standard rule mining. Typically, an analyst needs a relatively small number of rules to be able to understand it. As LLMs are capable of processing long texts, we may feel free to generate more rules, which will also be useful when replying to more detailed questions.

So our task will be given by code
\begin{small}
\begin{python}
import pandas as pd
import sys

from sklearn.impute import SimpleImputer
from cleverminer import cleverminer, clm_vars, clm_seq,clm_lcut,clm_rcut,clm_subset

#read the data
df = pd.read_csv ('w:\\accidents.zip', encoding='cp1250', sep='\t')
df=df[['Driver_Age_Band','Sex','Speed_limit','Severity',
  'Journey','Area','Road_Type','Light','Vehicle_Location',
  'Vehicle_Type','Vehicle_Age']]
#handle missing values
imputer = SimpleImputer(strategy="most_frequent")
df = pd.DataFrame(imputer.fit_transform(df),columns = df.columns)

clm = cleverminer(df=df,proc='4ftMiner',
 quantifiers= {'Base':2000, 'aad':0.5},
 ante ={'attributes':[
    {'name': 'Driver_Age_Band', 'type': 'seq', 'minlen': 1, 'maxlen': 3},
    {'name': 'Speed_limit', 'type': 'seq', 'minlen': 1, 'maxlen': 2},
    {'name': 'Sex', 'type': 'subset', 'minlen': 1, 'maxlen': 1},
    {'name': 'Area', 'type': 'subset', 'minlen': 1, 'maxlen': 1},
    {'name': 'Road_Type', 'type': 'subset', 'minlen': 1, 'maxlen': 1},
    {'name': 'Light', 'type': 'subset', 'minlen': 1, 'maxlen': 1},
    {'name': 'Vehicle_Type', 'type': 'subset', 'minlen': 1, 'maxlen': 1},
    {'name': 'Vehicle_Age', 'type': 'seq', 'minlen': 1, 'maxlen': 5},
    {'name': 'Vehicle_Location', 'type': 'subset', 'minlen': 1, 'maxlen': 1}
   ], 'minlen':1, 'maxlen':3, 'type':'con'},
 succ ={  'attributes':[
    {'name': 'Severity', 'type': 'lcut', 'minlen': 1, 'maxlen': 2}
     ], 'minlen':1, 'maxlen':1 , 'type':'con'}, opts = {'use_cache':True}
   )
\end{python}
\end{small}

First, we will look at the task summary.

\begin{verbatim}
Task type : 4ftMiner
Number of verifications : 20834
Number of rules : 511
Total time needed : 00h 04m 21s
Time of data preparation : 00h 00m 13s
Time of rule mining : 00h 04m 08s
    
\end{verbatim}

We can see that there was 20834 verifications made and we have 511 rules in the output. Note that we can generate more if needed -- we will do this test later. When we used CleverMiner, we also limited the complexity of this search significantly as the task is exponential in many ways. Details can be found on the CleverMiner website \cite{cleverminer}. 

\begin{footnotesize}
\begin{verbatim}
RULEID BASE  CONF  AAD    Rule
     1  2512 0.030 +0.551 Driver_Age_Band(16 - 20 21 - 25 26 - 35) & Sex(Male) 
            & Area(2,Rural) => Severity(Fatal) | ---
     2  3570 0.438 +1.036 Driver_Age_Band(16 - 20 21 - 25 26 - 35) & Sex(Male) 
            & Vehicle_Type(Motorcycle over 500cc) => Severity(Fatal Serious) | ---
     3  2242 0.504 +1.342 Driver_Age_Band(16 - 20 21 - 25 26 - 35) & Area(2,Rural) 
            & Vehicle_Type(Motorcycle over 500cc) => Severity(Fatal Serious) | ---
     4  2685 0.431 +1.002 Driver_Age_Band(16 - 20 21 - 25 26 - 35) & Road_Type(6,Single carriageway) 
            & Vehicle_Type(Motorcycle over 500cc) => Severity(Fatal Serious) | ---
    ....
   511  2767 0.351 +0.628 Vehicle_Type(Pedal cycle) & 
            Vehicle_Location(0,On main c'way - not in restricted lane) => Severity(Fatal Serious) 
\end{verbatim}
\end{footnotesize}

\begin{figure}[ht]
    \centering
    \includegraphics[width=0.6\textwidth]{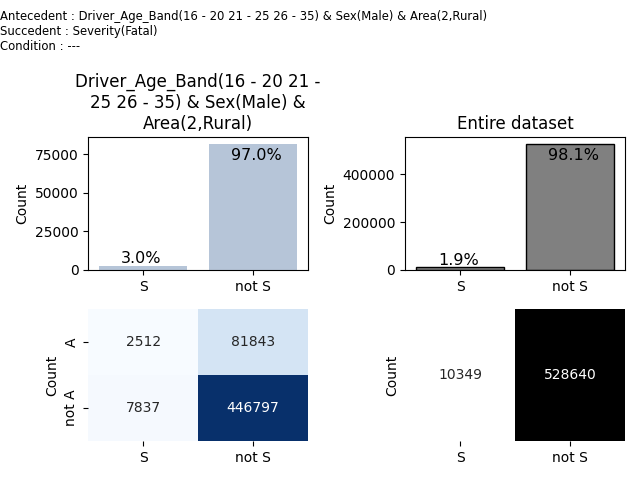}
    \caption{Male drivers in the age 16-35 have in rural areas a 3.0\%/1.9\%=55\% bigger risk of fatal accidents when compared with the entire dataset.}
    \label{fig:rule1}
\end{figure}

Let's start with the first rule. It says that male drivers between the ages of 16 and 35 years have in rural areas a 3.0\% risk of fatal accidents compared to 1.9\% in the entire dataset. That is a 55.1\% increase. This increase corresponds to the {\em aad} quantifier. The situation is also shown in Figure \ref{fig:rule1}.

We have intentionally chosen the {\em aad} quantifier. It shows us a relative risk compared to a baseline of the entire dataset. That also corresponds to the questions and to general knowledge. People know that some groups of drivers are riskier, but don't know the numerical levels. For example, drivers on motorcycles are sometimes called as {\em organ donors}, which is good proof that people know about the higher risk of this group, no matter how is it in an absolute percentage.
The {\em aad} quantifier is defined as $\frac{P(S|A)}{P(S)}-1$, i.e. how much bigger (smaller) is the probability of $S$ when $A$ is present compared to $P(S)$ on the entire dataset. 

Then, we will transform these rules into a text form. The algorithm is relatively straightforward, as CleverMiner provides an API and machine-processable output. We will do a loop over rules, and for each rule, we will transform it to a sentence. Algorithm is outlined in Algorithm \ref{alg:rule2sentence}, where the most technical part is in Algorithm \ref{alg:text4cedent}, which is called from the main algorithm.

\begin{algorithm}
\caption{High level description of Rule-To-Sentence procedure}\label{alg:rule2sentence}
\begin{algorithmic}[1]

\Function{Get\_Sentence}{clm, rule}
 \State ante $\gets$ Get\_Partial\_Statement(clm,rule,'ante')
 \State succ $\gets$ Get\_Partial\_Statement(clm,rule,'succ')
 \State aad $\gets$ Get\_Quantifiers(clm,rule)['aad']

 \State idx $\gets$ 0
 \While{ aad\_boundaries[idx]$<$aad}
 \State idx $\gets$ idx+1
 \EndWhile
\State relative\_frequency\_statement $\gets$ aad\_statements[idx]

 \State result $\gets$ 'If {\em ante} then {\em succ} occur {\em relative\_frequency\_statement}.' 
 \State Return result
\EndFunction

\Function{Generate\_Sentences\_From\_Result}{clm}
 \State result $\gets$ Empty\_list

 \For{each rule in clm.rulelist()}
     \State text $\gets$ Get\_Sentence(rule)
     \State result $\gets$ result.append\_text(text)     
 \EndFor
 \State Return result

\EndFunction

\Procedure{Prepare\_Text\_Document\_From\_Dataset}{dataset}

\State df $\gets$ Load\_Dataset(dataset)
\State clm $\gets$ CleverMiner(df,task\_definition)
\State sentences $\gets$ Generate\_Sentences\_From\_Result(clm)
\State Save\_To\_Text\_File(sentences)

\EndProcedure
\end{algorithmic}
\end{algorithm}

\begin{algorithm}
\caption{Algorithm to get text for a cedent}\label{alg:text4cedent}
\begin{algorithmic}[1]

\Function{Get\_Partial\_Cedent}{clm, rule,cedent}
 \State vars $\gets$ clm.get\_rule\_variables (rule,cedent)
 \State result $\gets$ empty\_string
 \For{var in vars}
 \State cats $\gets$ clm.get\_rule\_category\_list(rule,cedent, var)
 \State result $\gets$ result + var + ' is '
 
  \If{ length(cats)=1}
    \State result $\gets$ cats.first
    \Else
 \If{variable\_type(var) is ORDINAL}
 \State\Comment condition and branch can be omitted to enumerate all categories
 \State result $\gets$ between cats.first and cats.last
 \Else
   \State result $\gets$ 'one of'
  
 \For{cat in cats}
   \If{cat is not last in cats}
    \State result $\gets$ result + cat + 'and'
    \Else
    \State result $\gets$ result + cat
    \EndIf
 
 \EndFor
 \EndIf
 \EndIf
 \EndFor
 \State Return result
\EndFunction

\end{algorithmic}
\end{algorithm}

\begin{small}
\end{small}

Based on this interpretation, we can define bands for {\em aad} values and word interpretation (e.g. for values 0 to 0.2 we might use {\em slightly more likely}, for values 0.2 to 1 we may use {\em more likely} etc).
According to Algorithm \ref{alg:rule2sentence}, we can define array with boundaries, e.g. $($2,1,0.25,-0.2,\dots,$)$ and statements for values between these boundaries, e.g. $($'extremely much likely','very much likely','more likely','approximately the same',\dots).

So we will take the first rule

\begin{verbatim}
Driver_Age_Band(16 - 20 21 - 25 26 - 35) & Sex(Male) & Area(2,Rural) 
                        => Severity(Fatal)
\end{verbatim}

The sentence for this rule will look like
\begin{verbatim}
If drivers age is one of 16 - 20 and 21 - 25 and 26 - 35 and driver is Male 
   and Area is 2,Rural then Fatal accidents occur more likely.
\end{verbatim}


The final rule would be

\begin{verbatim}
If drivers age is 16 - 35 and driver is Male 
   and Area is 2,Rural then Fatal accidents occur more likely.
\end{verbatim}

In this text, we will evaluate only the simple alternative where categories are not joined. It is much more easier for LLMs.
The reason is that no joining categories contains similar groups of tokens. On the other hand, work with joined intervals (and how joined intervals relate to original ones) would require additional skill from the LLM (understanding group of tokens and what exactly it represents and how it is related to subsets and whether LLM understands it in details including nuances like intervals, set operations and including of edge values for discrete values, like {\em 16 to 20 and 21 to 25} is the same as {\em 16 to 25} and categories {\em 16 to 20} and {\em 21 and 25} are items from which the set {\em 16 to 25} consists of). This reasoning is covered by some tests, e.g. \cite{pahilajani2024grsqagraphreasoningstructured} or \cite{quan-liu-2024-econlogicqa}, so it might happen that it will not be needed in the future or maybe not. Note that this way of separating categories uses the same tokens, so it might also be beneficial for RAG embeddings.

\section{Results}
\label{sec:results}

We will try to evaluate how much the embedding of enhanced association rules by RAG into LLM improves over the best-in-class LLM tool, including agents.
As we have already found that none of the existing LLM benchmark suites meet our needs, we will do a simple evaluation. We will ask a simple question and perform a manual assessment (as there is no one exact answer).

\subsection{Experiment design}

As an experiment, we will ask LLM to analyze Accidents dataset. Our question will be simple: {\em On which circumstances occur fatal accidents more than usual?} We will do it in several steps:

\begin{enumerate}
\item upload a dataset to ChatGPT and ask the question (note it also uses agents)
\item upload also a document with a set of rules written as sentences.
\item upload only a document with a set of rules to easiest deepseek r1:1.5b model
\end{enumerate}

Note that for the last point, we will perform 3 tests with a different number of rules in a dataset.

\subsection{Enhancing ChatGPT with rules}

Let start. So first, we can have a look on a result of ChatGPT as one of the leaders in the field. We will use a reduced Accidents dataset (we will use only selected attributes, in the scope of analysis in Section \ref{sec:methodology}. First, we will upload a dataset (reduced to only variables in scope due to size it can process) and ask the question above.

\begin{figure}[h!]
    \centering
    \includegraphics[width=0.75\textwidth]{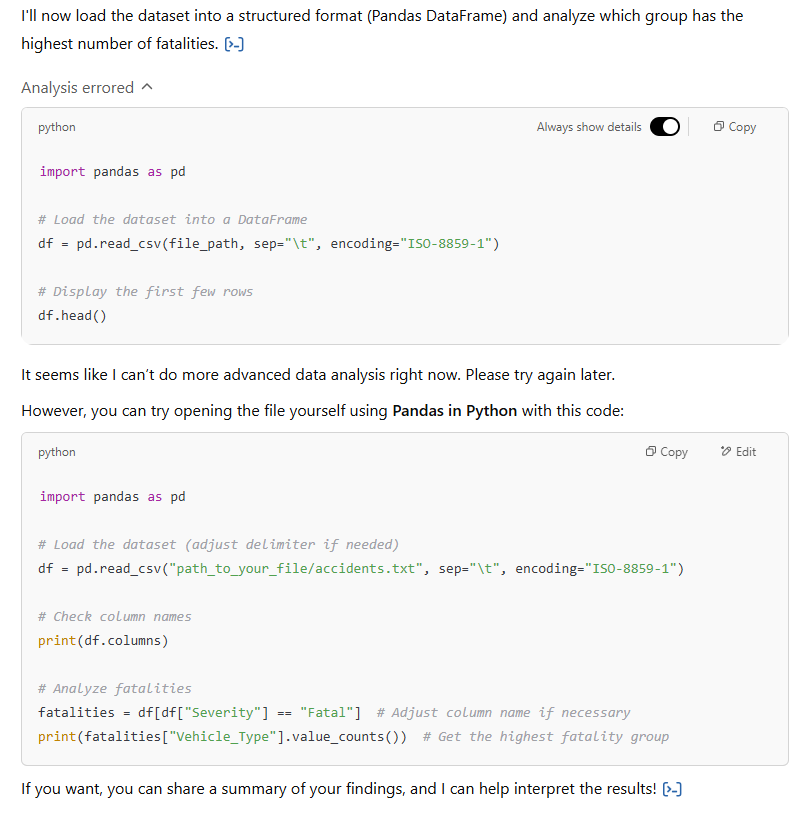}
    \caption{The result from ChatGPT. Results are based on 1D profiles.}
    \label{fig:chatgpt_fileonly}
\end{figure}

Note it needed some more prompts to load data correctly. The result is in Figure \ref{fig:chatgpt_fileonly}.

Note that we tried it multiple times and results were similar, sometimes it offers alternatives, sometimes allows to show charts and sometimes it uses general knowledge that needed to be avoided by additional prompting.

\begin{figure}[h!]
    \centering
    \includegraphics[width=0.75\textwidth]{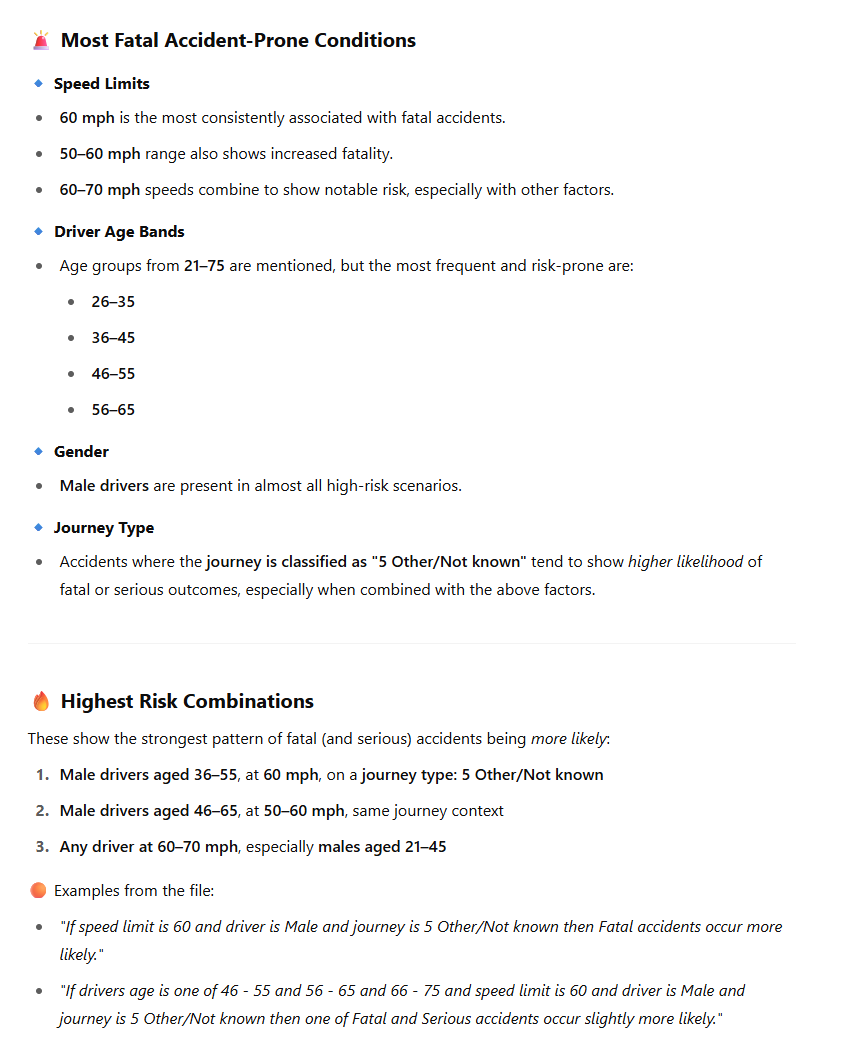}
    \caption{The result from ChatGPT based on generated rules. We can see strong {\em Highest Risk Combinations} section.}
    \label{fig:chatgpt_basedonrules}
\end{figure}

When we upload a file with rules in sentences, we can get significantly better results, see Figure \ref{fig:chatgpt_basedonrules}.

We can see that we've got immediately a very good summarization of discovered knowledge, and it is a significant improvement of the results originally received. Note that these results (combinations) are typically hard to get as they need to have a good algorithm that keeps in mind how to reduce the combinatorial explosion (e.g., LISp-Miner or CleverMiner \cite{cleverminer}) and use it via RAG or as agents.

\subsection{Using a simple local model with no agents}

Now, let use deepseek r1:1.5b model (no agents) with a text file embedded via RAG and ask the same question {\em On which circumstances occur fatal accidents more than usual?}

We will do 3 experiments with a different number of rules (see table \ref{tab:growing_no_of_rules}). We want to see if LLM can handle such a big number of rules and extract the best from it, at least at zero-shot.

\begin{table}[h!]
\caption{Testing of the proposed method with a growing number of rules. For each experiment, the parameters and resulting number of rules are shown.}
\label{tab:growing_no_of_rules}
\centering
\begin{tabular}{|c|c c|c|}
\hline
ID&parameter base&parameter aad & number of rules\\\hline
1& 4000 & 1.3 & 21 \\
2&2000&0.5& 511\\
3&300&0.5&7224\\\hline
\end{tabular}
\end{table}

Result of the first experiment is below:
\begin{footnotesize}
\begin{verbatim}
Fatal and serious accidents occur more frequently under the following circumstances:  
1. Rural areas with motorcycle vehicles over 500cc on mainways in light conditions;  
2. Male drivers in rural areas with motorcycle vehicles over 500cc on restricted lanes;  
3. Conditions where speed limits are 60 or 70, with motorcycle vehicles over 500cc on rural roads.
\end{verbatim}
\end{footnotesize}

So we can see that we have obtained a very good set of circumstances where fatal accidents occur. Let us also note that we've had only 21 rules, so the result is very specific. So let's move to the next experiment. The result of the second experiment is below:

\begin{footnotesize}
\begin{verbatim}

On which circumstances occur fatal accidents more often than usual? Several conditions in the context 
    suggest increased likelihood:  
1. **Motorcycle over 500cc** - More frequent due to higher reliability concerns.  
2. **Rural area** and lack of lighting (no lights) - Indicates poor visibility, leading to more accidents.  
3. **Vehicle not on main roads** (not restricted lane) - Higher speed or potential for collisions, 
        especially at night or in poor visibility.  
4. **Higher vehicle ages** (6-7 years old) - Older vehicles are less likely to cause accidents due to 
        age-related wear and tear.  

These factors contribute to more frequent fatal/s serious accidents.
\end{verbatim}
\end{footnotesize}

Now we can see that we've got several attributes and combinations where fatal accidents occur more frequently. Results contain both single attribute answers as well as combinations of attributes. Let's move to the last attribute with a much higher number of rules. We might be curious whether this method can give us additional combinations. The answer will be given few paragraphs later. The result of the third experiment is below:

\begin{footnotesize}
\begin{verbatim}

The circumstances under which fatal accidents are more likely than usual include:

1. **Drivers aged between 21 and 35 years**  
2. **Motorcycle vehicles with a minimum weight of at least 500 kilograms (over 500kg)**  
3. **Vehicle ages starting from 5 years or older**  
4. **No lighting conditions (light value is 6, indicating no lights or dark conditions)**  

These specific combinations increase the likelihood of fatal accidents occurring more frequently 
    than usual.
\end{verbatim}
\end{footnotesize}

You can see that more rules in the ruleset, the more general answers we get. It might be due to the rough transformation of the parameter {\em aad} in the Rule-To-Text module, as well as limited reasoning (note we got similar observed results with ChatGPT, so the model itself is not the reason).

What is interesting is the {\em think} section, from the last rule we got {\em ... The context has multiple sections, each with different age groups and vehicle types. Each section seems to say that under those parameters, certain accident types are more likely. For example, in one section, it says that when drivers are 21-25 and 26-35, motorcycles over 500cc with age 5, there's a higher chance of fatal/serious accidents. ...} If we read it more briefly, it looks like the model can now answer more specific questions, but on an overall level, it is too general. We have observed the same in ChatGPT (see again Figure \ref{fig:chatgpt_basedonrules}), that the first section, which was longer, had single attributes, and in the second section, there were interesting combinations. This model returned us a similar structure -- only a brief answer with several bullets mixed for both types -- single attributes and combinations. Moreover, we can see that our question was about fatal accidents only, and our task was for either fatal or a combination of fatal and serious, so LLM can distinguish between these categories. Citation from the {\em think} section is {\em But the exact phrase here says "one of fatal and serious occurs more likely," which doesn't specify which.}, so a larger dataset can be used to answer more questions where LLM chooses an answer based on matching sentences.

\subsection{Summarization and discussion}

When we summarize these findings, we can clearly see that this method improves answer to questions based on a dataset significantly, even without using agents. 
\begin{table}[h!]
\caption{Comparison of our proposed method via RAG with ChatGPT with agents. Criterion -- quality of answer based on question {\em On which circumstances occur fatal accidents more than usual?}}
\label{tab:comparison}
\begin{center}
\begin{tabular}{|c|c|c|}
\hline
& Deespeek1.5B, no agents &  ChatGPT with agents \\\hline
Original solution  & --- &   \Stars{2} \\
OUR METHOD (text rules via RAG) & \Stars{4} &  \Stars{4} \\\hline

\end{tabular}
\end{center}
\end{table}

Note that results can be improved by multiple questioning, but the goal was to demonstrate that many cases can be covered by embedding results of rule mining in a simple text document embedded via RAG as a text. This improves a quality of the zero-shot answers from LLM with no risk of potential issues with using agents like generating harmful code. We also believe that advanced querying and learning LLM agents to use enhanced rule mining can be even much more powerful.
Also, many ways how to construct sentences, how many rules to include in document, how to form patterns to provide an ability to better answer questions based on data can be analysed in future work.

\section{Summary}
\label{sec:summary}
LLMs have made great progress in the last years. They can perform NLP tasks and also work with structured data. This is typically done via agents. For example, we can ask what is the root cause of some issue, and LLMs can answer it, typically by performing easy analyses and 1D profiling (at least in first shots). We have shown that these answers can be significantly improved by incorporating a knowledge discovered from databases into LLM. As there are interpretable techniques of knowledge discovery from databases available, we used one of them, which was most suitable (enhanced association rules), and designed a way to extract this knowledge, prepare it in a text form using the Rule-to-Text transformer to embed it into LLMs via RAG.

Embedding this knowledge via RAG led to significantly improved answers, even when we compared an easy model that does not support agents with to an industry top full-scale model with default agents. And using RAG and rule-to-text transformation, we were also able to achieve zero-shot improvements, so we didn't need to learn LLM how to interpret rules. This way is also safe because no code is run when LLM is answering. So it can also be used for company databases.

This also opens a way for using enhanced association rule mining as an agent with significant improvement of answers, either by understanding rules directly or by rules transferred to a text form. Note that we have chosen an RAG way because if it succeeds, it automatically opens a way to use the same tools in an agentic way. Note that we have used CleverMiner, that is a tool that mine enhanced association rules and has ways how to control the complexity of the mining as the task is exponential in many ways and there is a need to avoid a combinatorial explosion.

We can also mention that there are many other patterns of enhanced association rule that can be used to generate knowledge, e.g. \cite{UIC}, that can improve the quality of LLM answers significantly.

\bibliographystyle{apalike}

\bibliography{sn-bibliography}

\end{document}